\documentclass[10pt,twocolumn,letterpaper]{article}
\usepackage{authblk}
\usepackage[pagenumbers]{cvpr}
\newcommand{\CVPRPaperTitle}{DriftWAM: Faster World-Action Models via Action-Aware Video Drifting}

\usepackage{algorithm}
\usepackage{algpseudocode}
\usepackage{multirow}

\newcommand{\mv}{\mathcal{V}}
\newcommand{\ma}{\mathcal{A}}

\usepackage{microtype}

\definecolor{cvprblue}{rgb}{0.21,0.49,0.74}
\usepackage[pagebackref,breaklinks,colorlinks,allcolors=cvprblue]{hyperref}

\def\confName{CVPR}
\def\confYear{2027}
\title{\CVPRPaperTitle}

\author[1]{Minyang Li}
\author[2]{Siyu Xu}
\author[2]{Yunke Wang}
\author[1,3]{Ying-Cong Chen}
\author[2]{Chang Xu}
\author[4]{Daochang Liu\thanks{Corresponding author.}}
\affil[1]{The Hong Kong University of Science and Technology (Guangzhou)}
\affil[2]{The University of Sydney}
\affil[3]{The Hong Kong University of Science and Technology}
\affil[4]{The University of Western Australia}

\begin{document}
\raggedbottom
\maketitle

\begin{abstract}
World-action models (WAMs) unify future video prediction and action generation for robotic control, but their reliance on iterative denoising incurs substantial inference costs.
Distilling these models into few-step generators offers a promising route to efficient deployment, yet maintaining task performance under aggressive step reduction remains challenging.
Existing methods rely on point-wise trajectory consistency distillation, which pursues high visual fidelity while leaving downstream action information unexploited in optimizing video generation, and also lacks explicit distribution-level alignment.
To address this, we propose \textbf{DriftWAM}, a few-step WAM distillation framework that incorporates the actions induced by predicted videos into video learning.
Specifically, DriftWAM integrates \emph{action-aware video drifting}, which performs video distribution matching through a joint visual--action similarity kernel to steer sample evolution under the guidance of these induced actions, with \emph{action-guided video regularization}, which leverages the compatible teacher--student matches identified by the joint kernel to backpropagate action-level gradients directly into the video generator.
Evaluations on the RoboTwin 2.0 and LIBERO benchmarks, as well as in real-world robotic manipulation tasks, show that DriftWAM consistently outperforms prior consistency distillation methods across few-step settings at the same inference speed.
\href{https://minyangli.github.io/driftwam-project-page/}{Project page}.
\end{abstract}

\section{Introduction}
\label{sec:intro}

\begin{figure}[t]
    \centering
    \captionsetup{skip=10pt}
    \includegraphics[width=\columnwidth]{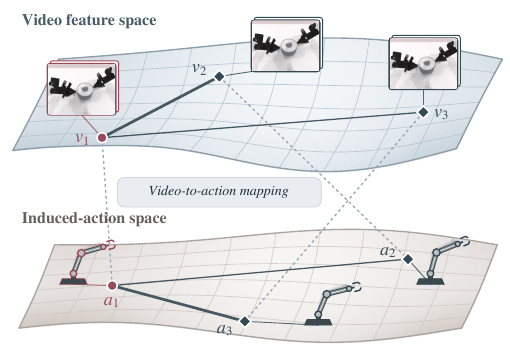}
    \caption{\textbf{Disagreement between visual and induced-action perspectives.}
    Given candidate videos generated under the same context, candidate $v_2$ is closer to the reference video $v_1$ than $v_3$ in video feature space, whereas $v_3$ induces an action $a_3$ closer to $a_1$ than $a_2$ in induced-action space.}
    \label{fig:teaser}
    \vspace{6pt}
\end{figure}

World-action models (WAMs) have emerged as a promising paradigm for embodied intelligence, unifying future video prediction and robot action generation within a generative framework~\cite{shen2025videovla,bi2026motus,li2026causal,kim2026cosmospolicy,ye2026world}.
By leveraging rich spatiotemporal priors from large-scale video pretraining~\cite{yang2025cogvideox,wan2025wan,nvidia2025cosmospredict2},
WAMs predict how the environment will evolve given current observations and task instructions, and generate the corresponding robot actions~\cite{du2023learning,feng2026vidar,pai2025mimicvideo,li2026causal}. %
This visual foresight grounds decision-making in physical dynamics, enabling strong generalization across complex robotic manipulation tasks.
Despite their effectiveness, WAMs incur substantial inference latency because future video synthesis and action prediction often rely on iterative diffusion or flow-matching sampling~\cite{ho2020ddpm,lipman2023flowmatching,chi2023diffusionpolicy}. %
As a result, generating each control decision typically requires multiple sequential denoising steps to produce high-dimensional video latents and the corresponding action sequence.
This computational burden stands in sharp conflict with the real-time reactivity required for closed-loop robotic deployment.
Few-step distillation~\cite{song2023consistency,prasad2024consistencypolicy,yin2024dmd,yin2024dmd2,zhang2026teacherfeature,akbari2026flashwam} %
therefore offers a natural optimization strategy, aiming to compress multi-step denoising into one or a few steps to reduce inference latency while preserving the capabilities of the full multi-step model.

Existing efforts toward few-step WAM distillation typically adapt consistency distillation~\cite{song2023consistency,luo2023lcm,wang2023videolcm} from visual synthesis to each modality branch.
For example, Flash-WAM~\cite{akbari2026flashwam} shows that high-dimensional, redundant video latents and low-dimensional, precision-critical actions operate under different noise regimes, so applying the same visual consistency objective to both branches can lead to severe performance degradation.
It therefore adopts an action consistency function whose gradient-scaling coefficient varies linearly with the noise level, while retaining the standard latent consistency model (LCM)~\cite{luo2023lcm} consistency function for the video branch.

Despite this progress, such modality-aware consistency distillation adaptations only tailor separate objectives to each data modality, while their video distillation does not explicitly consider how the generated videos affect downstream actions.
This omission matters, as video and action exhibit an inherent \emph{functional asymmetry} in WAMs.
In other words, we judge a WAM by whether its actions accomplish the task, while its predicted videos serve as intermediate representations that guide the generation of those actions.
Given this video--action functional asymmetry, we argue that video learning in WAMs should not focus solely on visual fidelity.
We further find that assessing the similarity between two videos from the perspective of visual appearance versus from the perspective of their impact on the actions leads to considerable disagreements (Fig.~\ref{fig:teaser}; see Appendix~\ref{sec:app_disagreement}).
Therefore, we believe that beyond pursuing visual fidelity, video learning in WAMs should also leverage the actions induced by predicted videos.

However, incorporating these actions should not compromise the original capability of generating visually faithful videos.
Thus, for few-step WAM distillation, we need a suitable approach that preserves the ability to pursue visual fidelity as much as possible, while offering a low-cost and effective interface for incorporating action information.
To this end, we draw inspiration from the drifting paradigm~\cite{deng2026drifting}, which shifts iterative distributional refinement from inference to training.
Drifting fulfills our needs for two reasons:
1) it governs sample attraction and repulsion directly through pairwise similarity kernels, naturally accommodating action-level signals;
2) it enables one-step inference and has shown strong promise in diffusion distillation~\cite{zhang2026teacherfeature} and world modeling~\cite{lu2026driftworld} for high-fidelity image and video generation.
Moreover, we find that, as a distribution matching method, drifting brings the added benefit of achieving better alignment between the student's and teacher's conditional video distributions (see Sec.~\ref{sec:mmd_analysis}).

Building on this foundation, we propose \textbf{DriftWAM}, a few-step WAM distillation framework that unites \emph{action-aware video drifting} and \emph{action-guided video regularization}.
Specifically, action-aware video drifting combines visual feature distances with the discrepancies between the actions induced by predicted videos into a joint similarity kernel, which reshapes sample evolution in feature space.
Furthermore, to complement this kernel-weighted feature evolution with explicit supervision on induced-action discrepancies, we reuse the compatible teacher--student pairs selected by the joint kernel to construct an action-guided video regularizer, through which action-level gradients are backpropagated directly into the video generator.

\begin{figure*}[t]
    \centering
    \includegraphics[width=\textwidth]{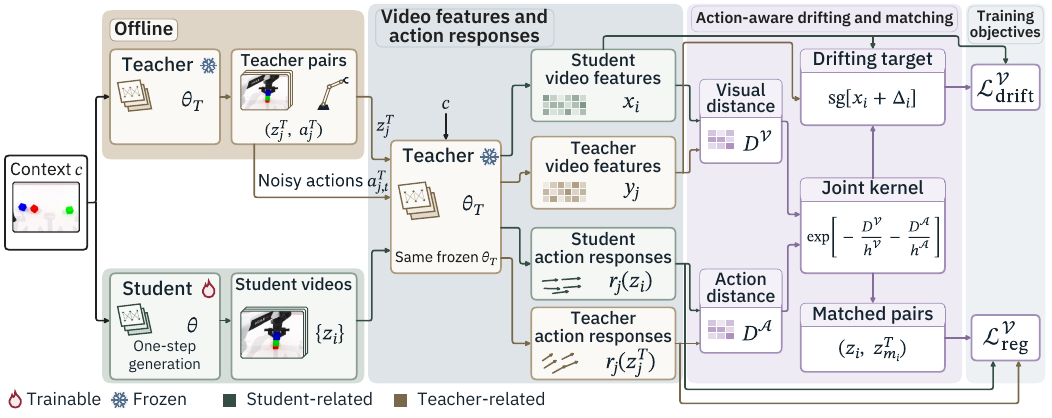}
    \caption{\textbf{Training Overview of the video branch in DriftWAM.}
    On the left, the teacher generates reference video--action pairs
    offline, while the student produces video candidates under the
    same context $c$.
    Sharing the same frozen parameters, the teacher in the middle extracts visual features and computes action responses
    for both student and teacher videos.
    The resulting visual and action-response distances feed the joint
    kernel, whose attraction and repulsion weights determine the targets for $\mathcal{L}^{\mv}_{\mathrm{drift}}$ in the upper branch.
    In parallel, the lower branch uses the same kernel to select
    student--teacher pairs for $\mathcal{L}^{\mv}_{\mathrm{reg}}$, which aligns
    the paired videos' action responses and directly backpropagates
    action-level gradients to the video generator.}
    \label{fig:training_overview}
\end{figure*}

In summary, our main contributions are as follows:
\begin{itemize}
    \item Based on the inherent video--action functional asymmetry in WAMs and our finding of the misalignment between visual and induced-action similarities among candidate videos, we argue that few-step video distillation should account for the downstream actions induced by predicted videos alongside visual fidelity.
    \item We develop \textbf{DriftWAM}, in which \emph{action-aware video drifting} injects induced-action similarity into the drifting kernel to guide video distribution matching, while \emph{action-guided video regularization} exploits the resulting compatible teacher--student pairs to directly supervise the video generator via action-level gradients.
    \item Extensive evaluations on the RoboTwin 2.0~\cite{chen2026robotwin} and LIBERO~\cite{liu2023libero} benchmarks, as well as in real-world robotic manipulation tasks, demonstrate that DriftWAM maintains strong control performance even when inference is compressed into one or a few denoising steps, delivering higher task success rates than state-of-the-art consistency distillation under identical inference budgets.
\end{itemize}

\section{Related work}
\label{sec:related_work}

\subsection{World-action models}
\label{sec:related_wam}
Leveraging spatiotemporal priors from pretrained video generators~\cite{wan2025wan,du2023learning,wu2024gr1}, world-action models (WAMs) unify future visual prediction and action generation via diffusion or flow matching~\cite{shen2025videovla,bi2026motus,li2026causal}.
Existing WAMs that synthesize explicit visual futures either denoise video and action tokens jointly~\cite{guo2024pad,bi2026motus,ye2026world} or follow a cascade formulation that conditions action generation on predicted video latents~\cite{li2026causal,pai2025mimicvideo}.
To mitigate the latency of multi-step sampling~\cite{ye2026world,yuan2026fastwam}, prior works bypass test-time video generation~\cite{yuan2026fastwam,ye2026gigaworldpolicy}, predict futures in compact latent spaces~\cite{chen2026lawam}, adopt lightweight backbones with sparse tokens~\cite{li2026efficientwam}, or cache computation across inference chunks~\cite{zhao2026c3ache}.
Orthogonal to architectural or test-time modifications, we accelerate cascade WAMs while preserving explicit video-to-action inference through few-step distillation.

\subsection{Few-step distillation}
\label{sec:related_distillation}
Few-step distillation for diffusion and flow-matching models broadly follows two paradigms: trajectory-following methods such as consistency distillation~\cite{song2023consistency,luo2023lcm,wang2023videolcm}, which enforce point-wise agreement along denoising trajectories, and distribution-matching methods, which align output distributions via auxiliary score critics~\cite{yin2024dmd,yin2024dmd2} or critic-free drifting~\cite{deng2026drifting,zhang2026teacherfeature}.
Extending step distillation to WAMs, Flash-WAM~\cite{akbari2026flashwam} introduces modality-specific consistency functions to account for the distinct noise regimes of video and action branches.
Concurrent works also explore adaptive step budgets~\cite{wang2026anystepwam}, interaction-centric visual guidance~\cite{lyu2026dido}, or deterministic direct prediction (DirectWAM~\cite{zheng2026directwam}).
While DirectWAM matches induced actions to demonstrated actions through a separately trained one-step action model, DriftWAM directly reuses the frozen teacher for action-aware video distribution matching and compatible teacher video selection.

\section{Preliminaries}
\label{sec:preliminaries}

\subsection{Problem formulation}
\label{sec:problem_formulation}
We study few-step distillation of cascade WAMs, which generate robot actions conditioned on predicted visual futures~\cite{li2026causal,pai2025mimicvideo}.
Let $c$ denote the available context, including visual observations, previous actions, and a task instruction.
A model with parameters $\theta$ first samples a future video-latent sequence $z$ given $c$ and then generates an action sequence $a$ conditioned on both $z$ and $c$.
For brevity, we refer to these video-latent sequences as videos.
The resulting joint distribution factorizes as
\begin{equation}
    p_{\theta}(z,a \mid c)
    =
    p_{\theta}(z \mid c)\,
    p_{\theta}(a \mid z,c).
    \label{eq:wam_factorization}
\end{equation}
The video branch provides an imagined evolution of the scene, while the action branch determines robot commands conditioned on this future prediction and the observed context.

Both branches generate samples through iterative flow-based denoising with $N^{\mv}$ video steps and $N^{\ma}$ action steps. We denote this inference setting by $N^{\mv}\mathrm{v}/N^{\ma}\mathrm{a}$ (\eg, $1\mathrm{v}/1\mathrm{a}$ for one step per branch).
While a pretrained teacher with parameters $\theta_T$ achieves strong manipulation performance using multi-step sampling, directly reducing $N^{\mv}$ and $N^{\ma}$ at test time to accelerate inference severely degrades task performance.
Our goal is to distill the multi-step teacher $\theta_T$ into a student model $\theta$ that preserves the teacher's capabilities under a few-step inference budget.

\subsection{One-step generation via drifting}
\label{sec:one_step_drifting}
Drifting~\cite{deng2026drifting} trains a one-step generator to match a target distribution by evolving generated samples along a kernel-defined drift field during training.
Each generated sample experiences attraction toward target samples and repulsion from other generated samples, and the generator is updated by regressing toward the resulting drifted targets.

Given context $c$, a one-step generator $g_{\theta}$ maps noise $\epsilon_i\sim\mathcal{N}(0,I)$ to generated samples $z_i=g_{\theta}(\epsilon_i,c)$ ($i\in\{1,\dots,K\}$), which are compared against target samples $z_j$ ($j\in\{1,\dots,M\}$) drawn from the target distribution.
A fixed feature extractor $\phi$ yields features $x_i=\phi(z_i)$ and $y_j=\phi(z_j)$, and the Euclidean distance $D(x,y)=\|x-y\|_2$ defines a kernel $k(x,y)=\exp[-D(x,y)/h]$ with bandwidth $h>0$.
The normalized attraction and repulsion weights, $w^+_{ij}=k(x_i,y_j)/\sum_{j'=1}^{M}k(x_i,y_{j'})$ and $w^-_{i\ell}=k(x_i,x_\ell)/\sum_{\ell'=1,\,\ell'\ne i}^{K}k(x_i,x_{\ell'})$ ($\ell\ne i$), define the drift as follows:
\begin{equation}
    \Delta_i
    = \sum_{j=1}^{M} w^+_{ij}(y_j-x_i)
    - \sum_{\ell=1,\,\ell\ne i}^{K} w^-_{i\ell}(x_\ell-x_i).
    \label{eq:video_drift}
\end{equation}
Regressing generated features toward the drifted targets yields
\begin{equation}
    \mathcal{L}_{\mathrm{drift}}
    = \mathbb{E}\!\left[
        \left\|x_i-\operatorname{sg}[x_i+\Delta_i]\right\|_2^2
    \right],
    \label{eq:drifting_loss}
\end{equation}
where $\operatorname{sg}$ denotes the stop-gradient operator.

\section{Method}
\label{sec:method}\label{sec:driftwam}
In this section, we first describe how action information guides video distribution matching in Sec.~\ref{sec:action_aware_video_drifting}. We then introduce complementary action-level supervision for video generation in Sec.~\ref{sec:action_response_video_regularization}.
Section~\ref{sec:overall_training} describes the complete training procedure and presents the overall objective combining video and action supervision.
The training pipeline of the video branch is illustrated in Fig.~\ref{fig:training_overview}.

\subsection{Action-aware video drifting}
\label{sec:action_aware_video_drifting}
In video distillation via drifting, the teacher-generated videos can be treated as target samples, so that the one-step student learns to match the teacher's conditional video distribution.
For each context $c$, we denote the student videos by $\{z_i\}_{i=1}^{K}$ and the teacher videos by $\{z_j^T\}_{j=1}^{M}$.

\paragraph{Visual distance.}
Following Teacher-Feature Drifting~\cite{zhang2026teacherfeature}, we use the frozen teacher as the visual feature extractor $\phi^{\mv}$ to reuse representations learned for video prediction without introducing an additional encoder.
For a video $z$ under context $c$, its visual features are extracted by perturbing $z$ at a fixed noise level $t_f$ and passing it through the frozen teacher:
\begin{equation}
    \phi^{\mv}(z)
    = H_{\theta_T}^{(q)}
    \!\left((1-t_f)z+t_f\eta,\;t_f;\;c\right),
    \label{eq:video_feature_extraction}
\end{equation}
where $H_{\theta_T}^{(q)}$ denotes the hidden output of the $q$-th Transformer block in the teacher's video branch, and Gaussian noise $\eta\sim\mathcal{N}(0,I)$ is shared across candidate videos under the same context.
For any two candidate videos $z$ and $z'$, their visual distance is computed between the corresponding features $x=\phi^{\mv}(z)$ and $x'=\phi^{\mv}(z')$ as
\begin{equation}
    D^{\mv}(x,x')=\|x-x'\|_2,
\end{equation}
which applies to both student--teacher and student--student comparisons.

Yet visual feature distances alone need not reflect how differences between predicted videos affect action prediction.
We therefore incorporate action distances into the drifting kernel so that video learning accounts for the actions induced by predicted videos.

\paragraph{Action distance.}
For the action distance, a direct approach is to compare the action sequences that each video induces under the frozen teacher, as this assesses how each video influences downstream action generation under a common, well-trained reference policy.
However, generating these sequences requires costly multi-step denoising, so we instead compare flow-velocity predictions from the teacher's action branch at a shared noisy action input, which provides an effective single-step proxy for the similarity between fully denoised action sequences (see Appendix~\ref{sec:app_proxy_validation}).
Specifically, each teacher video $z_j^T$ is paired with an action sequence $a_j^T$ generated by the teacher conditioned on $(z_j^T,c)$.
We form its noisy version as $a_{j,t}^T=(1-t)a_j^T+t\epsilon$,
where the uniformly drawn timestep $t$ and Gaussian noise $\epsilon\sim\mathcal{N}(0,I)$ are shared across the compared videos.
Let $u_{\theta_T}^{\ma}$ denote the frozen teacher's action flow velocity field.
We then compute
\begin{equation}
    \begin{aligned}
        r_j(z) &= u_{\theta_T}^{\ma}(a_{j,t}^T,t;z,c), \\
        D^{\ma}_j(z,z') &= \left\|r_j(z)-r_j(z')\right\|_2^2,
    \end{aligned}
    \label{eq:action_response_distance}
\end{equation}
where $r_j(z)$ denotes the teacher's \emph{action response} to video $z$, and $D_j^{\ma}(z,z')$ measures the distance between the action responses to $z$ and $z'$.
The subscript $j$ identifies the teacher pair providing the shared noisy action input $a_{j,t}^T$.

\paragraph{Teacher--student attraction.}\hspace{-0.5em}
Combining visual and action-response distances gives the teacher--student attraction kernel
\begin{equation}
    k^+_{ij}
    = \exp\!\left[
        -\frac{D^{\mv}(x_i,y_j)}{h^{\mv}}
        -\frac{D^{\ma}_j(z_i,z_j^T)}{h^{\ma}}
    \right],
    \label{eq:action_aware_attraction}
\end{equation}
where $h^{\mv},h^{\ma}>0$ are the corresponding bandwidths.
Here, $x_i=\phi^{\mv}(z_i)$ and $y_j=\phi^{\mv}(z_j^T)$.
Under this joint visual--action criterion, we select $m_i=\arg\max_j k^+_{ij}$ and pair each student video $z_i$ with its most compatible teacher video $z_{m_i}^T$, resulting in $(z_i,z_{m_i}^T)$.

\paragraph{Student--student repulsion.}
In teacher--student attraction, the reference video $z_j^T$ determines the noisy action input $a_{j,t}^T$ used for comparison.
We extend this rule to student--student repulsion by using $a_{m_\ell,t}^T$ from the teacher matched to the reference student video $z_\ell$ (see Appendix~\ref{sec:equilibrium}).
Evaluating both $z_i$ and $z_\ell$ at this shared input gives
\begin{equation}
    k^-_{i\ell}
    = \exp\!\left[
        -\frac{D^{\mv}(x_i,x_\ell)}{h^{\mv}}
        -\frac{D^{\ma}_{m_\ell}(z_i,z_\ell)}{h^{\ma}}
    \right],
    \qquad \ell\ne i.
    \label{eq:action_aware_kernel}
\end{equation}

Normalizing $k^+_{ij}$ over teacher samples and $k^-_{i\ell}$ over the other students gives the action-aware attraction weights $w^+_{ij}$ and repulsion weights $w^-_{i\ell}$ in Eq.~\eqref{eq:video_drift}.
We substitute the resulting action-aware drift into the regression objective of Eq.~\eqref{eq:drifting_loss} and denote this video drifting loss by $\mathcal{L}^{\mv}_{\mathrm{drift}}$.

\subsection{Action-guided video regularization}
\label{sec:action_response_video_regularization}
Although action-aware video drifting uses action information to determine interaction weights, its updates are constructed in video feature space.
To complement this mechanism, we introduce an action-guided video regularizer that directly backpropagates action-level gradients into the video generator.
A teacher video selected solely by visual similarity may look close to the student prediction yet induce a different control behavior, making it a less suitable reference for this direct supervision.
We therefore reuse the student--teacher pair $(z_i,z_{m_i}^T)$ selected by the joint kernel, which considers both visual content and induced-action similarity to identify a more compatible teacher target.
We then minimize the discrepancy between the paired videos' action responses at the shared noisy action input $a_{m_i,t}^T$, giving
\begin{equation}
    \mathcal{L}^{\mv}_{\mathrm{reg}}
    = \mathbb{E}\!\left[
        \left\|r_{m_i}(z_i)
        - r_{m_i}(z_{m_i}^{T})\right\|_2^2
    \right].
    \label{eq:video_regularization}
\end{equation}

\subsection{Overall training}
\label{sec:overall_training}

\paragraph{Video branch training.}
As illustrated in Fig.~\ref{fig:training_overview}, video training uses $M$ teacher video--action pairs $\{(z_j^T,a_j^T)\}_{j=1}^{M}$ under context $c$.
To avoid repeatedly running multi-step teacher sampling during training, these teacher pairs are generated once offline to form a reusable dataset.
For each $c$, the student generates $K$ video candidates $\{z_i\}_{i=1}^{K}$ in one step from independent noise.
The frozen teacher then extracts visual features and computes action responses for both student and teacher videos.
The resulting visual and action-response distances are used in the joint kernel to obtain attraction and repulsion weights and form student--teacher pairs $(z_i,z_{m_i}^T)$.
Using these weights and pairs, we train the video branch with the action-aware drifting loss $\mathcal{L}^{\mv}_{\mathrm{drift}}$ and the action-guided video regularization loss $\mathcal{L}^{\mv}_{\mathrm{reg}}$.

\paragraph{Action branch training.}
Alongside these video objectives, consistency distillation~\cite{song2023consistency} is applied to the action branch using demonstration video--action pairs $(z,a)$ under context $c$.
The noisy action input is formed as $a_t=(1-t)a+t\epsilon$, where $\epsilon\sim\mathcal{N}(0,I)$.
Following Flash-WAM~\cite{akbari2026flashwam}, we use the student's action flow velocity field $u_{\theta}^{\ma}$ to define the action consistency function
$f_{\theta}^{\ma}(a_t,t;z,c)=a_t-t\,u_{\theta}^{\ma}(a_t,t;z,c)$,
which predicts the clean action and whose velocity coefficient scales linearly with $t$.
For a lower-noise timestep $s<t$, the frozen teacher produces
$\widetilde{a}_s=a_t+(s-t)u_{\theta_T}^{\ma}(a_t,t;z,c)$.
The action consistency loss is given by
\begin{equation}
    \mathcal{L}^{\ma}_{\mathrm{CD}}
    = \mathbb{E}\!\left[
        d\!\left(
            f_{\theta}^{\ma}(a_t,t;z,c),
            f_{\bar{\theta}}^{\ma}(\widetilde{a}_s,s;z,c)
        \right)
    \right],
    \label{eq:action_cd}
\end{equation}
where $\bar{\theta}$ denotes an exponential moving average (EMA) of the student parameters and $d$ is a discrepancy function.
The additional flow-matching regularizer~\cite{lipman2023flowmatching} is given by
\begin{equation}
    \mathcal{L}^{\ma}_{\mathrm{FM}}
    = \mathbb{E}\!\left[
        \left\|
            u_{\theta}^{\ma}(a_t,t;z,c)-(\epsilon-a)
        \right\|_2^2
    \right].
    \label{eq:action_fm}
\end{equation}

All four objectives contribute to the optimization of the student model $\theta$ through
\begin{equation}
    \mathcal{L}
    = \mathcal{L}^{\mv}_{\mathrm{drift}}
    + \lambda_{\mathrm{reg}}\mathcal{L}^{\mv}_{\mathrm{reg}}
    + \lambda_{\mathrm{CD}}\mathcal{L}^{\ma}_{\mathrm{CD}}
    + \lambda_{\mathrm{FM}}\mathcal{L}^{\ma}_{\mathrm{FM}}.
    \label{eq:joint_objective}
\end{equation}

\section{Experiments}
\label{sec:experiments}

\begin{table*}[t]
  \centering
  \caption{RoboTwin 2.0 success rates (\%) and inference speedup relative to the LingBot-VA teacher.
  $^*$Results reproduced by us.
  Best success rates within each inference setting are bolded.}
  \label{tab:robotwin_main_results}
  \resizebox{\textwidth}{!}{%
  \begin{tabular}{lcccccccccccc}
    \toprule
    & & & \multicolumn{2}{c}{Horizon $=1$}
          & \multicolumn{2}{c}{Horizon $=2$}
          & \multicolumn{2}{c}{Horizon $=3$}
          & \multicolumn{3}{c}{Average (50 tasks)}
          & \\
    \cmidrule(lr){4-5}
    \cmidrule(lr){6-7}
    \cmidrule(lr){8-9}
    \cmidrule(lr){10-12}
    Method & $N^{\mv}$ & $N^{\ma}$
           & Clean & Rand.
           & Clean & Rand.
           & Clean & Rand.
           & Clean & Rand. & Avg.
           & Speedup \\
    \midrule
    LingBot-VA$^*$ (teacher)
      & 25 & 50 & 92.87 & 93.13 & 86.80 & 87.40 & 88.00 & 88.80 & 90.56 & 90.98 & 90.77 & 1.00$\times$ \\
    \midrule
    Flash-WAM$^*$
      & 1 & 2 & 91.40 & 88.60 & 83.20 & 75.67 & \textbf{73.00} & 65.60 & 87.10 & 82.42 & 84.76 & 17.15$\times$ \\
    DriftWAM (ours)
      & 1 & 2 & \textbf{91.73} & \textbf{89.80} & \textbf{85.07} & \textbf{82.27} & 71.60 & \textbf{73.80} & \textbf{87.72} & \textbf{85.94} & \textbf{86.83} & 17.15$\times$ \\
    \midrule
    Flash-WAM$^*$
      & 1 & 1 & 85.93 & 82.40 & 71.73 & 67.00 & 53.60 & 44.40 & 78.44 & 73.98 & 76.21 & 21.46$\times$ \\
    DriftWAM (ours)
      & 1 & 1 & \textbf{87.23} & \textbf{86.00} & \textbf{81.40} & \textbf{77.40} & \textbf{62.20} & \textbf{65.80} & \textbf{82.98} & \textbf{81.40} & \textbf{82.19} & 21.46$\times$ \\
    \bottomrule
  \end{tabular}%
  }
\end{table*}

\begin{table*}[t]
  \centering
  \small
  \begin{minipage}[t]{0.57\textwidth}
    \centering
    \caption{LIBERO success rates (\%).
    $^*$Results reproduced by us.
    $^\dagger$Results directly reported in Flash-WAM~\cite{akbari2026flashwam}.
    Best results within each setting are bolded.}
    \label{tab:libero_results}
    \setlength{\tabcolsep}{3.5pt}
    \begin{tabular*}{\textwidth}{@{\extracolsep{\fill}}lcccccc}
      \toprule
      Method & $N^{\mv}$/$N^{\ma}$ & Spatial & Object & Goal & Long & Avg. \\
      \midrule
      LingBot-VA$^*$ (teacher) & 20/50 & 98.1 & 99.2 & 98.3 & 98.1 & 98.4 \\
      \midrule
      Flash-WAM$^\dagger$ & 1/2 & 97.0 & 92.8 & 96.4 & \textbf{98.0} & 96.1 \\
      DriftWAM (ours) & 1/2 & \textbf{97.2} & \textbf{99.1} & \textbf{97.5} & 96.1 & \textbf{97.5} \\
      \midrule
      Flash-WAM$^\dagger$ & 1/1 & \textbf{96.0} & 92.6 & 96.0 & \textbf{95.8} & 95.1 \\
      DriftWAM (ours) & 1/1 & 95.9 & \textbf{99.0} & \textbf{96.8} & \textbf{95.8} & \textbf{96.9} \\
      \bottomrule
    \end{tabular*}
  \end{minipage}\hfill
  \begin{minipage}[t]{0.40\textwidth}
    \centering
    \caption{Conditional video distribution discrepancy (teacher: $N^{\mv}=25$; students: $N^{\mv}=1$).
    \emph{Teacher--Teacher} compares two independent teacher sample groups as a reference. Its finite-sample estimate may be slightly negative.}
    \label{tab:video_distribution_mmd}
    \renewcommand{\arraystretch}{0.94}
    \setlength{\tabcolsep}{4pt}
    \resizebox{0.96\linewidth}{!}{%
    \begin{tabular*}{\linewidth}{@{\extracolsep{\fill}}lc}
      \toprule
      Distribution Comparison & $\mathrm{MMD}^2 \downarrow$ \\
      \midrule
      Teacher--Teacher (reference) & $-0.0018$ \\
      \midrule
      Teacher--Flash-WAM~\cite{akbari2026flashwam} & $0.9272$ \\
      Teacher--Vanilla Video Drifting & $0.4167$ \\
      \textbf{Teacher--DriftWAM (ours)} & $\mathbf{0.3402}$ \\
      \bottomrule
    \end{tabular*}%
    }
  \end{minipage}
\end{table*}

\subsection{Experimental setup}
\label{sec:exp_setup}

\paragraph{Base model.}
We instantiate DriftWAM on LingBot-VA~\cite{li2026causal}, a cascade WAM that first predicts future video and then conditions action generation on the predicted videos.
On RoboTwin 2.0, we initialize the student from the consistency-distilled Flash-WAM checkpoint~\cite{akbari2026flashwam} by default, and further ablate initializing directly from the pretrained LingBot-VA teacher in Sec.~\ref{sec:ablation} for a comparison of the two methods as standalone distillation frameworks.
On LIBERO, since no Flash-WAM checkpoint is publicly available to date, we initialize the student directly from the LingBot-VA teacher.
For training, we use $M=K=4$ teacher and student candidates per context. On RoboTwin 2.0, training takes about 24 hours on 4 NVIDIA A800 GPUs (see Appendix~\ref{sec:supp-implementation} for full training details).

\paragraph{Benchmarks.}
We evaluate DriftWAM on two simulation benchmarks and a real-world robotic setup:
(i) \textbf{RoboTwin 2.0}~\cite{chen2026robotwin}, comprising 50 bimanual manipulation tasks across three horizon lengths ($\text{Horizon}=1, 2, 3$) under Clean and Randomized splits (100 rollouts per task per split), where the Randomized split perturbs scene clutter, lighting, background textures, and tabletop heights;
(ii) \textbf{LIBERO}~\cite{liu2023libero}, covering the Spatial, Object, Goal, and Long suites (10 tasks each); and
(iii) \textbf{Real-world manipulation} on a 6-DoF AgileX Piper arm (with third-person and wrist-mounted RGB cameras) across three tasks covering precision placement, pick-and-place, and multi-stage stacking (20 rollouts per task under matched randomized object placements), where LingBot-VA is fine-tuned on 93 teleoperated demonstrations (40,852 frames) before distilling Flash-WAM and DriftWAM.

\paragraph{Evaluation metrics.}
Beyond task success rates, we measure per-chunk inference latency in simulation on a single NVIDIA RTX 4090 GPU to report speedup relative to the LingBot-VA teacher, and report replanning time for real-world evaluation.
To verify that DriftWAM achieves distribution-level video refinement, we also evaluate the squared maximum mean discrepancy ($\mathrm{MMD}^2$)~\cite{gretton2012kernel} with an LPIPS~\cite{zhang2018lpips} perceptual kernel between teacher and student video distributions across 30 conditioning histories on RoboTwin 2.0 (see Appendix~\ref{sec:app_mmd} for details).

\paragraph{Baselines.}
We compare DriftWAM against the multi-step teacher LingBot-VA~\cite{li2026causal} and the state-of-the-art consistency distillation baseline Flash-WAM~\cite{akbari2026flashwam}.
In simulation, the teacher uses its default multi-step schedule ($25\mathrm{v}/50\mathrm{a}$ on RoboTwin 2.0 and $20\mathrm{v}/50\mathrm{a}$ on LIBERO), while the distilled models are evaluated under two few-step budgets: $1\mathrm{v}/2\mathrm{a}$ and $1\mathrm{v}/1\mathrm{a}$.
In real-world evaluation (Table~\ref{tab:real_world}), we report the teacher at its released deployment budget ($3\mathrm{v}/10\mathrm{a}$) and compare DriftWAM, Flash-WAM, and direct step-reduced LingBot-VA under the $1\mathrm{v}/2\mathrm{a}$ budget.

\begin{figure*}[t]
  \centering
  \includegraphics[width=\textwidth]{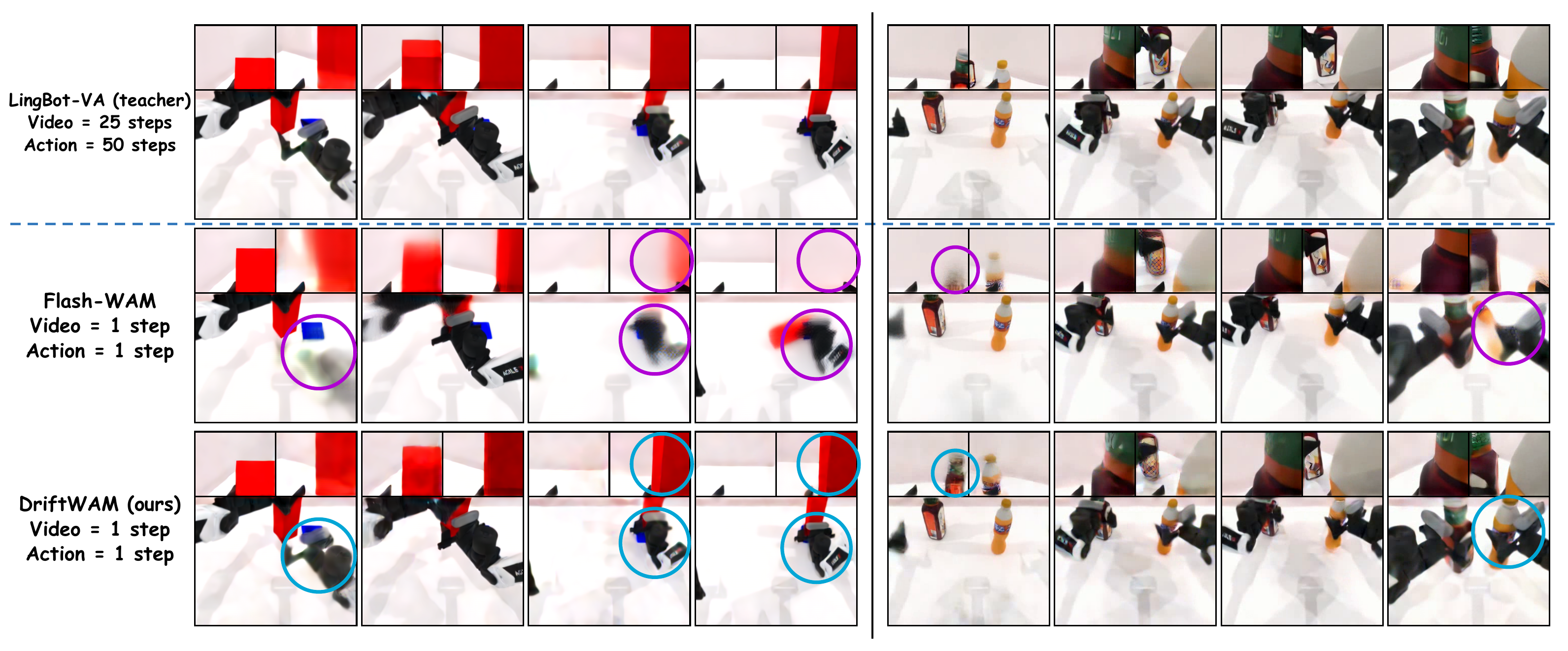}
  \caption{Qualitative comparison on RoboTwin 2.0 \texttt{handover\_block} (left) and \texttt{pick\_diverse\_bottles} (right) under one-step ($1\mathrm{v}/1\mathrm{a}$) inference.}
  \label{fig:qualitative}
\end{figure*}

\subsection{Main results}
\label{sec:main_results}

\paragraph{RoboTwin 2.0.}
Table~\ref{tab:robotwin_main_results} reports the results on RoboTwin 2.0.
DriftWAM outperforms Flash-WAM across both step budgets at the same inference speedup (see Appendix~\ref{sec:app_latency}), with the performance margin widening substantially as inference is compressed from $1\mathrm{v}/2\mathrm{a}$ to one-step $1\mathrm{v}/1\mathrm{a}$.
This difference arises because a two-step action branch still retains capacity to iteratively refine noisy predictions, whereas single-step action generation has no room for iterative correction and becomes highly sensitive to the quality of the predicted video, amplifying the benefit of stronger visual conditioning from DriftWAM.
Further examining the breakdown across evaluation splits and task horizons shows that these gains are concentrated on the Randomized split and scale sharply with task horizon, reaching $+10.40$ and $+21.40$ percentage points on $\text{Horizon}=2$ and $\text{Horizon}=3$ under $1\mathrm{v}/1\mathrm{a}$.
This pattern stems from how our action-aware kernel and regularizer guide video generation to focus on action-relevant regions, which improves robustness under domain randomization and mitigates visual-control errors from compounding across multi-stage tasks.

\paragraph{LIBERO.}
Table~\ref{tab:libero_results} summarizes the results on LIBERO.
Although we compare against the numbers reported in Flash-WAM~\cite{akbari2026flashwam} and initialize directly from the LingBot-VA teacher due to the absence of a released Flash-WAM checkpoint to date, the results exhibit a consistent trend with RoboTwin 2.0: DriftWAM outperforms Flash-WAM across both inference budgets, with a larger overall margin under the one-step ($1\mathrm{v}/1\mathrm{a}$) setting. Reducing the action branch from two steps to one causes only a modest drop in DriftWAM's average success, showing that the distilled student closely preserves the multi-step teacher's capability even under the smallest tested inference budget.

\paragraph{Qualitative analysis.}
Figure~\ref{fig:qualitative} compares two examples of video predictions on RoboTwin 2.0 \texttt{handover\_block} and \texttt{pick\_diverse\_bottles} under one-step ($1\mathrm{v}/1\mathrm{a}$) inference against the LingBot-VA teacher ($25\mathrm{v}/50\mathrm{a}$).
The multi-step teacher maintains clear object identity and gripper geometry across both tasks.
Under one-step video generation, Flash-WAM exhibits severe visual degradation as the rollout progresses: during critical grasping and handover phases, both the target objects (the red block and bottles) and the grippers undergo structural collapse and blurring, depriving the action branch of reliable spatial cues.
In contrast, DriftWAM preserves sharp object boundaries, intact gripper structure, and coherent arm trajectories that closely track the teacher across frames.
This visual comparison aligns with our quantitative findings in Table~\ref{tab:robotwin_main_results}, confirming that DriftWAM effectively prevents structural degradation in action-relevant regions under aggressive step compression.

\begin{figure*}[t]
  \centering
  \includegraphics[width=\textwidth]{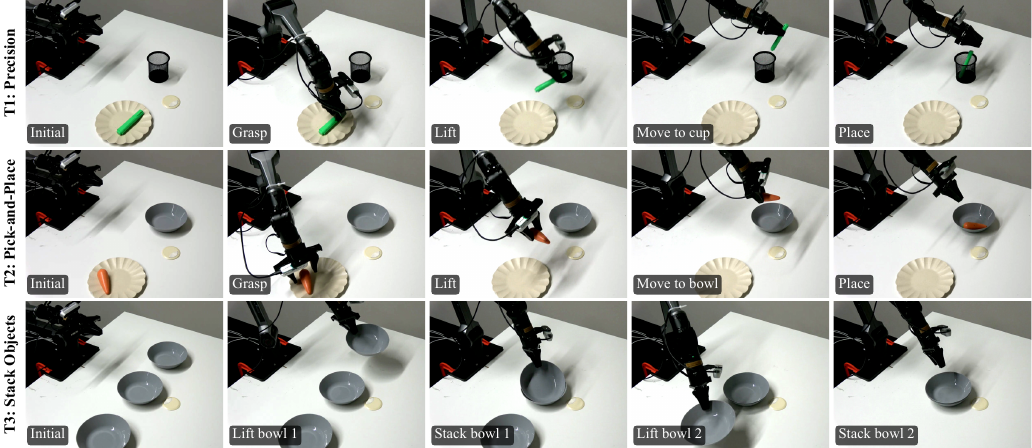}
  \caption{Keyframes from the three real-world tasks on the AgileX Piper arm, with time progressing from left to right.
  T1: Pick up the pen and place into cup; T2: Place the carrot into the bowl; T3: Stack the three bowls.}
  \label{fig:real_world_tasks}
\end{figure*}

\paragraph{Real-world evaluation.}
Figure~\ref{fig:real_world_tasks} illustrates the three real-world tasks and Table~\ref{tab:real_world} reports the evaluation results.
Directly reducing the LingBot-VA teacher from $3\mathrm{v}/10\mathrm{a}$ to $1\mathrm{v}/2\mathrm{a}$ without distillation causes severe degradation across all tasks.
Under the same $1\mathrm{v}/2\mathrm{a}$ budget, DriftWAM achieves the highest average success rate, outperforming Flash-WAM on both pick-and-place (T2) and multi-stage bowl stacking (T3) and closely approaching the multi-step teacher. This contrast with direct step reduction demonstrates DriftWAM's ability to recover much of the teacher's control performance at a comparable replanning time.

\subsection{Video distribution alignment analysis}
\label{sec:mmd_analysis}

Table~\ref{tab:video_distribution_mmd} evaluates conditional video distribution alignment on RoboTwin 2.0 (see Appendix~\ref{sec:app_mmd} for details).
Compared with the near-zero Teacher--Teacher reference, Flash-WAM exhibits a considerable distributional gap, showing that point-wise trajectory consistency alone is insufficient to closely align the conditional video distribution under one-step video generation.
Replacing trajectory consistency with Vanilla Video Drifting cuts this discrepancy by more than half, and DriftWAM reduces it even further.
Together with the control improvements in Table~\ref{tab:ablation_robotwin}, this result suggests that incorporating action information into video drifting further improves both task performance and alignment with the teacher's video distribution.

\begin{table}[t]
  \centering
  \small
  \captionsetup{skip=8pt}
  \renewcommand{\arraystretch}{1.15}
  \caption{Ablation study of DriftWAM components on RoboTwin 2.0 (success rates, \%) under the one-step ($1\mathrm{v}/1\mathrm{a}$) setting.
  All variants are initialized from Flash-WAM.
  $^*$Results reproduced by us.}
  \label{tab:ablation_robotwin}
  \setlength{\tabcolsep}{5.8pt}
  \begin{tabular*}{\columnwidth}{@{\extracolsep{\fill}}lccc}
    \toprule
    Method Variant & Clean & Rand. & Avg. \\
    \midrule
    Flash-WAM$^*$ & 78.44 & 73.98 & 76.21 \\
    Vanilla Video Drifting & 81.36 & 76.84 & 79.10 \\
    Vanilla + Action-Aware Kernel & 81.78 & 79.36 & 80.57 \\
    Vanilla + Action-Guided Reg. ($\mathcal{L}^{\mv}_{\mathrm{reg}}$) & 81.08 & 77.58 & 79.33 \\
    \textbf{DriftWAM (Full)} & \textbf{82.98} & \textbf{81.40} & \textbf{82.19} \\
    \bottomrule
  \end{tabular*}
\end{table}

\begin{table}[t]
  \centering
  \small
  \captionsetup{skip=8pt}
  \renewcommand{\arraystretch}{1.15}
  \caption{Ablation study on student initialization on RoboTwin 2.0 (success rates, \%) under the one-step ($1\mathrm{v}/1\mathrm{a}$) setting.
  $^*$Results reproduced by us.}
  \label{tab:ablation_init}
  \begin{tabular*}{\columnwidth}{@{\extracolsep{\fill}}lccc}
    \toprule
    Method & Clean & Rand. & Average \\
    \midrule
    Flash-WAM$^*$ & 78.44 & 73.98 & 76.21 \\
    DriftWAM (LingBot-VA init.) & 80.82 & 74.72 & 77.77 \\
    \textbf{DriftWAM (Flash-WAM init.)} & \textbf{82.98} & \textbf{81.40} & \textbf{82.19} \\
    \bottomrule
  \end{tabular*}
\end{table}

\subsection{Ablation study}
\label{sec:ablation}

\begin{table*}[t]
  \centering
  \small
  \captionsetup{skip=4pt}
  \renewcommand{\arraystretch}{0.96}
  \caption{Real-world evaluation on an AgileX Piper arm (replanning time, successful rollouts out of 20, and average success rate).
  Best success rates under the $1\mathrm{v}/2\mathrm{a}$ budget are bolded.
  T1: ``Pick up the pen and place into cup''; T2: ``Place the carrot into the bowl''; T3: ``Stack the three bowls''.}
  \label{tab:real_world}
  \setlength{\tabcolsep}{3.5pt}
  \begin{tabular*}{\textwidth}{@{\extracolsep{\fill}}llccccc}
    \toprule
    Method & $N^{\mv}$/$N^{\ma}$ & Replan Time (s) & T1 (pen $\to$ cup) & T2 (carrot $\to$ bowl) & T3 (stack bowls) & Average \\
    \midrule
    LingBot-VA (teacher)                 & 3/10 & 2.36 & 12/20 & 7/20 & 13/20 & 53.3\% \\
    \midrule
    LingBot-VA (reduced denoising steps) & 1/2  & 1.50 & 6/20 & 3/20 & 6/20 & 25.0\% \\
    Flash-WAM                            & 1/2  & 1.42 & \textbf{10/20} & 4/20 & 10/20 & 40.0\% \\
    \textbf{DriftWAM (ours)}             & 1/2  & 1.43 & 9/20 & \textbf{6/20} & \textbf{12/20} & \textbf{45.0\%} \\
    \bottomrule
  \end{tabular*}
\end{table*}

\paragraph{Component analysis.}
Table~\ref{tab:ablation_robotwin} ablates the core components of DriftWAM on RoboTwin 2.0 under the one-step ($1\mathrm{v}/1\mathrm{a}$) setting (see Appendix~\ref{sec:app_ablation_1v2a} for the $1\mathrm{v}/2\mathrm{a}$ ablation).
Consistent with the $\mathrm{MMD}^2$ reduction in Table~\ref{tab:video_distribution_mmd}, vanilla video drifting already improves downstream control over Flash-WAM.
Incorporating the action-aware kernel brings a further gain, especially on the Randomized split.
Moreover, the action-guided video regularizer ($\mathcal{L}^{\mv}_{\mathrm{reg}}$) relies on this action-aware kernel to be effective.
Adding $\mathcal{L}^{\mv}_{\mathrm{reg}}$ to vanilla video drifting with teacher matches selected by visual distance alone yields almost no improvement, whereas combining it with the action-aware kernel in full DriftWAM achieves the best performance across both splits.
This synergy arises because $\mathcal{L}^{\mv}_{\mathrm{reg}}$ supervises each student video toward its matched teacher reference.
Selecting this reference by visual distance alone often pairs the student with teachers that are visually close but less consistent in their induced actions, weakening the regularizer's supervision, whereas our joint kernel selects more action-compatible teacher matches.

\paragraph{Student initialization.}
Table~\ref{tab:ablation_init} compares different student initializations on RoboTwin 2.0 under the one-step ($1\mathrm{v}/1\mathrm{a}$) setting.
When initialized directly from the pretrained LingBot-VA teacher as a standalone distillation framework, DriftWAM achieves a modest improvement over Flash-WAM.
Between the two initializations, starting from Flash-WAM achieves stronger performance because, under one-step video generation, a student initialized from LingBot-VA produces noisier videos than the consistency-distilled Flash-WAM checkpoint, making the action guidance computed by the frozen teacher's action branch less reliable in early training.

\section{Conclusion}
\label{sec:conclusion}

We presented DriftWAM, a few-step distillation framework for cascade world-action models that addresses the video--action functional asymmetry overlooked by trajectory consistency distillation while mitigating conditional video distribution discrepancy.
DriftWAM couples action-aware video drifting with action-guided video regularization. It evaluates candidate videos with the teacher's action branch to incorporate action information into video distribution matching, while directly backpropagating action-level gradients through the resulting compatible teacher--student matched pairs.
Experiments on RoboTwin 2.0 and LIBERO demonstrate that DriftWAM is effective both as a standalone distillation method and as a refinement over consistency-distilled checkpoints, yielding pronounced gains under one-step inference, visual perturbations, and long-horizon manipulation without additional test-time overhead.
Real-world experiments on an AgileX Piper arm further show that, at the $1\mathrm{v}/2\mathrm{a}$ budget, DriftWAM achieves the highest average success rate among the compared methods and retains $84.4\%$ of the teacher's average success.
However, while our formulation may also be applicable to unified joint-denoising WAMs, this work only validates its motivation and effectiveness on cascade WAMs, and multi-sample candidate generation still incurs extra training cost. Future work will therefore empirically investigate unified WAMs and improve training-time sample efficiency.

{\small
\bibliographystyle{ieeenat_fullname}
\bibliography{main}
}

\clearpage
\appendix
\maketitlesupplementary

\raggedbottom

\section{Visual--action similarity disagreement}
\label{sec:app_disagreement}

We examine whether visually closer predicted futures induce
more similar actions.
We measure visual distance $d_{\mathrm{vis}}$ using RMS differences between
$\ell_2$-normalized DINOv3 patch features~\cite{simeoni2025dinov3}, and action distance $d_{\mathrm{full}}$
using the MSE between action sequences generated by the frozen teacher through 50 denoising steps.
Action distances are computed in normalized action space
and averaged over shared initial noise draws.

\paragraph{Evaluation protocol.}
We use 30 conditioning histories from 10 RoboTwin 2.0 tasks,
with eight future videos generated by LingBot-VA per history.
For each group size $n\in\{3,4,5\}$, we enumerate all
groups of $n$ videos within each history and randomly
select one video in each group as the reference
$z_{\mathrm{ref}}$.
This yields $30\binom{8}{n}$ comparisons for each group size.
Among the remaining candidates $\{z_j\}_{j=1}^{n-1}$,
the indices of the visually nearest and action-nearest videos are given by
\begin{equation}
\begin{aligned}
    j_{\mathrm{vis}}
    &= \arg\min_j d_{\mathrm{vis}}(z_{\mathrm{ref}},z_j),\\
    j_{\mathrm{act}}
    &= \arg\min_j d_{\mathrm{full}}(z_{\mathrm{ref}},z_j).
\end{aligned}
\label{eq:app_disagreement_selection}
\end{equation}
We report the fraction of comparisons for which
$j_{\mathrm{vis}}\ne j_{\mathrm{act}}$.

\paragraph{Results.}
Table~\ref{tab:app_disagreement} shows substantial disagreement
between visual and action similarity, occurring in 41.37\%
of comparisons even with only two candidates per reference.
These results demonstrate that visually closer futures
do not necessarily induce more similar actions, motivating
the incorporation of action information into video learning.

\begin{table}[H]
    \centering
    \caption{\textbf{Visual--action similarity disagreement.}
    Each group contains one reference and $n-1$ candidates.}
    \label{tab:app_disagreement}
    \begin{tabular}{cccc}
        \toprule
        Group size $n$ & Comparisons & Disagreements & Rate \\
        \midrule
        3 & 1,680 & 695   & 41.37\% \\
        4 & 2,100 & 1,239 & 59.00\% \\
        5 & 1,680 & 1,122 & 66.79\% \\
        \bottomrule
    \end{tabular}
\end{table}

\section{Validation of the action-response proxy}
\label{sec:app_proxy_validation}

In Sec.~\ref{sec:action_aware_video_drifting}, DriftWAM uses
single-pass teacher flow-velocity predictions as an
action-response proxy.
Here, we examine whether this proxy serves as an effective
surrogate for full-denoising action similarity.

\subsection{Experiment 1: Action agreement of selected references}

\paragraph{Evaluation protocol.}
For each context $c$, we use $K=M=4$ student videos
$\{z_i\}_{i=1}^{K}$ and teacher videos
$\{z_j^T\}_{j=1}^{M}$.
Following the notation in the main paper, we compare
visual-only and joint selection:
\begin{equation}
    m_i^{\mathrm{vis}}
    = \arg\min_j D^{\mv}(x_i,y_j),
    \qquad
    m_i = \arg\max_j k^+_{ij}.
    \label{eq:app_proxy_selection}
\end{equation}
When the selections differ, we evaluate which selected
teacher video induces actions closer to those induced
by the student video.
All action sequences are generated by the same frozen
teacher through 50 denoising steps.
Action distance is measured by MSE in normalized action
space, averaged over three shared initial noise draws
sampled independently of proxy construction.

\paragraph{Setup and results.}
We use 30 conditioning histories from RoboTwin 2.0,
with student videos generated by the official Flash-WAM
checkpoint and teacher videos generated by LingBot-VA.
This yields 120 matching decisions.
Visual-only and joint selection choose different teacher
videos in 59 cases.
Among these disagreements, full-denoising action distances
favor the joint selection in 47 cases (79.66\%).

\subsection{Experiment 2: Pairwise action-distance ordering}

\paragraph{Evaluation protocol.}
We compare the relative ordering of teacher candidates
under the proxy and full-denoising action distances.
For student video $z_i$ and teacher candidate $z_j^T$,
let
$d^{\mathrm{proxy}}_{ij}=D_j^{\ma}(z_i,z_j^T)$
denote the action-response proxy distance, and let
$d^{\mathrm{full}}_{ij}$ denote the corresponding
full-denoising action distance.
For each student $i$, we consider every pair of distinct
teacher candidates $z_j^T$ and $z_{j'}^T$, with
$1\leq j<j'\leq M$.
The two distances agree on which teacher candidate
is closer to the student if
\begin{equation}
    \bigl(d^{\mathrm{proxy}}_{ij}
          -d^{\mathrm{proxy}}_{ij'}\bigr)
    \bigl(d^{\mathrm{full}}_{ij}
          -d^{\mathrm{full}}_{ij'}\bigr) > 0.
    \label{eq:app_proxy_ordering}
\end{equation}
We compute the full-denoising distances as in Experiment~1.

\paragraph{Setup and results.}
We use the same 30 contexts, with eight Flash-WAM
student videos and $M=4$ teacher candidates per context,
yielding 240 student queries.
For each student, comparing every pair among its four
teacher candidates gives $\binom{M}{2}=\binom{4}{2}=6$
ordering comparisons, or 1,440 comparisons in total.
The proxy and full-denoising distances agree in
1,210 comparisons (84.03\%).
This result supports the proxy's ability to preserve
the relative action similarity of teacher candidates
to a student future.

\section{Matched-Distribution Equilibrium}
\label{sec:equilibrium}

We analyze the video objectives in Secs.~\ref{sec:action_aware_video_drifting}--\ref{sec:action_response_video_regularization} for a fixed context and teacher reference set.
Let $P$ denote the uniform empirical distribution over the teacher
videos $\{z_j^T\}_{j=1}^M$ and $Q$ the student video distribution.

\subsection{Equilibrium of action-aware video drifting}
\label{sec:equilibrium_drifting}

Let $m(z)$ denote the teacher selected by the joint-kernel matching
rule in Sec.~\ref{sec:action_aware_video_drifting}, so that $m(z_i)=m_i$ for a student video $z_i$.
Define
\begin{equation}
    \begin{aligned}
        \kappa(z,z')=\exp\!\Big[
        &-\frac{
            D^{\mv}(\phi^{\mv}(z),\phi^{\mv}(z'))
        }{h^{\mv}}\\
        &-\frac{D_{m(z')}^{\ma}(z,z')}{h^{\ma}}
        \Big].
    \end{aligned}
    \label{eq:app_directed_affinity}
\end{equation}

Assume that the teacher visual features are pairwise distinct.
Under the joint-kernel matching rule, each teacher video's
affinity with itself is $1$, since both its visual and
induced-action distances to itself are zero.
Its affinity with any other teacher reference is strictly
smaller than $1$, because the visual distance is positive
and the induced-action distance is nonnegative.
Therefore, each teacher video selects itself:
\begin{equation}
    m(z_j^T)=j.
    \label{eq:app_self_matching}
\end{equation}
The self-matching property lets us use the same affinity $\kappa$ for
attraction to teacher references and repulsion from other student samples.

For a distribution $R$, define its affinity-weighted feature mean:
\begin{equation}
    \mu_R(z)=
    \frac{
        \mathbb{E}_{z'\sim R}
        [\kappa(z,z')\phi^{\mv}(z')]
    }{
        \mathbb{E}_{z'\sim R}[\kappa(z,z')]
    }.
    \label{eq:app_weighted_feature_mean}
\end{equation}

Recall that the drifting update in Sec.~\ref{sec:action_aware_video_drifting} operates on
feature displacements:
\[
    \Delta_i
    =
    \sum_j w_{ij}^{+}(y_j-x_i)
    -
    \sum_{\ell\ne i}w_{i\ell}^{-}(x_\ell-x_i),
\]
where $x_i=\phi^{\mv}(z_i)$ is the current student feature.
The first term attracts $x_i$ toward teacher features,
while subtracting the second term repels it from other
student features.
Since the weights in each term sum to one,
each weighted sum of displacements equals the weighted
mean of the reference features minus $x_i$.
For example,
\[
    \sum_j w_{ij}^{+}(y_j-x_i)
    =
    \sum_j w_{ij}^{+}y_j-x_i.
\]
Using the weighted means $\mu_P(z)$ and $\mu_Q(z)$ under $P$ and $Q$
gives the distribution-level drifting field
\begin{equation}
    \begin{aligned}
        \Delta(z)
        &=
        \big[\mu_P(z)-\phi^{\mv}(z)\big]
        -
        \big[\mu_Q(z)-\phi^{\mv}(z)\big]\\
        &=\mu_P(z)-\mu_Q(z),
    \end{aligned}
    \label{eq:app_population_drift}
\end{equation}
where the current feature $\phi^{\mv}(z)$ cancels between
the two terms.

\newcounter{appendixproposition}
\refstepcounter{appendixproposition}
\label{prop:app_matched_distribution}
\paragraph{Proposition \theappendixproposition{} (Matched-distribution equilibrium).}
If $Q=P$, then $\Delta(z)=0$ for every $z$.

\noindent\emph{Proof.}
Fix a query video $z$.
When $Q=P$, expectations under the two distributions are identical.
In particular,
\begin{equation}
    \begin{aligned}
        \mathbb{E}_{z'\sim P}
        [\kappa(z,z')\phi^{\mv}(z')]
        &=
        \mathbb{E}_{z'\sim Q}
        [\kappa(z,z')\phi^{\mv}(z')],\\
        \mathbb{E}_{z'\sim P}[\kappa(z,z')]
        &=
        \mathbb{E}_{z'\sim Q}[\kappa(z,z')].
    \end{aligned}
\end{equation}
Thus, both the numerators and denominators of the
affinity-weighted feature means coincide, giving
\begin{equation}
    \mu_P(z)=\mu_Q(z).
\end{equation}
Substituting this equality into
Eq.~\eqref{eq:app_population_drift} yields
\begin{equation}
    \Delta(z)=\mu_P(z)-\mu_Q(z)=0.
    \label{eq:app_matched_equilibrium}
\end{equation}
Since $z$ was arbitrary, the drifting field vanishes everywhere.
The regression target then equals the current visual feature,
so the corresponding drifting loss is zero.
\hfill$\square$

\subsection{Regularization at the matched distribution}
\label{sec:equilibrium_regularization}

Write the per-video regularization loss in Sec.~\ref{sec:action_response_video_regularization} as
\begin{equation}
    \ell_{\mathrm{reg}}^{\mv}(z)
    =
    \|r_{m(z)}(z)-r_{m(z)}(z_{m(z)}^T)\|_2^2,
    \label{eq:app_regularization_residual}
\end{equation}
with $\mathcal{L}_{\mathrm{reg}}^{\mv}(Q)
=\mathbb{E}_{z\sim Q}[\ell_{\mathrm{reg}}^{\mv}(z)]$.

\paragraph{Corollary 1 (Zero regularization loss at the matched distribution).}
The action-guided video regularizer vanishes at $Q=P$.

\noindent\emph{Proof.}
At $Q=P$, every student video is drawn from the teacher reference set.
Each teacher reference $z_j^T$ selects itself by Eq.~\eqref{eq:app_self_matching}, giving
\begin{equation}
    \ell_{\mathrm{reg}}^{\mv}(z_j^T)
    =\|r_j(z_j^T)-r_j(z_j^T)\|_2^2
    =0.
    \label{eq:app_zero_residual}
\end{equation}
Averaging over the teacher references gives
$\mathcal{L}_{\mathrm{reg}}^{\mv}(P)=0$.
Together with Proposition~\ref{prop:app_matched_distribution}, this shows that both video losses vanish at $Q=P$.
\hfill$\square$

\section{Video distribution alignment analysis}
\label{sec:app_mmd}

We examine how closely a few-step student aligns with the multi-step teacher's conditional video distribution using the squared maximum mean discrepancy ($\mathrm{MMD}^2$)~\cite{gretton2012kernel}.
The results are in the main paper.

\subsection{Formulation of Conditional Video \texorpdfstring{$\mathrm{MMD}^2$}{MMD Squared}}
\label{sec:app_mmd_formulation}

For a fixed conditioning history $c$, let $P(\cdot\mid c)$ and $Q(\cdot\mid c)$ denote two conditional future video distributions induced through random initial Gaussian noise.

Let $d(V,W)$ denote the LPIPS~\cite{zhang2018lpips} distance between generated videos $V$ and $W$, averaged over corresponding future frames and camera views.
Specifically, for videos consisting of $L$ predicted future frames across $U$ camera views, where $V_{u,\tau}$ denotes the frame at view $u\in\{1,\dots,U\}$ and future step $\tau\in\{1,\dots,L\}$,
\begin{equation}
    d(V,W)
    =
    \frac{1}{UL}\sum_{u=1}^{U}\sum_{\tau=1}^{L}
    d_{\mathrm{LPIPS}}\!\left(V_{u,\tau},\,W_{u,\tau}\right).
    \label{eq:app_video_lpips}
\end{equation}
We define the perceptual kernel
\begin{equation}
    k(V,W)
    =
    \exp\!\left(-\frac{d(V,W)}{h}\right),
    \label{eq:app_perceptual_kernel}
\end{equation}
where $h>0$ is a teacher-derived bandwidth.
Let $\psi$ denote its feature map into the reproducing kernel Hilbert space (RKHS) $\mathcal{H}$, satisfying $k(V,W)=\langle\psi(V),\psi(W)\rangle_{\mathcal{H}}$.
The two conditional distributions are represented by their kernel mean embeddings:
\begin{equation}
\begin{aligned}
    \nu_P
    &=
    \mathbb{E}_{V\sim P(\cdot\mid c)}[\psi(V)],\\
    \nu_Q
    &=
    \mathbb{E}_{W\sim Q(\cdot\mid c)}[\psi(W)].
\end{aligned}
\label{eq:app_kme}
\end{equation}
Their squared MMD is
\begin{equation}
    \mathrm{MMD}^2(P,Q)
    =
    \left\|\nu_P-\nu_Q\right\|_{\mathcal{H}}^{2}.
    \label{eq:conditional_video_mmd}
\end{equation}
This kernel formulation captures distributional differences beyond ordinary feature means: MMD vanishes if and only if the distributions coincide in the chosen perceptual representation.

Expanding Eq.~\eqref{eq:conditional_video_mmd} gives
\begin{equation}
\begin{aligned}
    \mathrm{MMD}^2(P,Q)
    ={}&
    \mathbb{E}_{V,V'\sim P(\cdot\mid c)}
    [k(V,V')]
    \\
    &+
    \mathbb{E}_{W,W'\sim Q(\cdot\mid c)}
    [k(W,W')]
    \\
    &-
    2\mathbb{E}_{\substack{
        V\sim P(\cdot\mid c)\\
        W\sim Q(\cdot\mid c)}}
    [k(V,W)],
\end{aligned}
\label{eq:video_mmd_expectation}
\end{equation}
where all draws in each expectation are independent.
Given $M$ samples $\{V_m\}_{m=1}^{M}\sim P(\cdot\mid c)$ and $K$ samples $\{W_n\}_{n=1}^{K}\sim Q(\cdot\mid c)$, we estimate this quantity by excluding self-pairs from the within-distribution terms~\cite{gretton2012kernel}:
\begin{equation}
\begin{aligned}
    \widehat{\mathrm{MMD}}^2(P,Q)
    ={}&
    \frac{1}{M(M-1)}
    \sum_{m=1}^{M}\sum_{m'\ne m} k(V_m,V_{m'})
    \\
    &+
    \frac{1}{K(K-1)}
    \sum_{n=1}^{K}\sum_{n'\ne n} k(W_n,W_{n'})
    \\
    &-
    \frac{2}{MK}
    \sum_{m=1}^{M}\sum_{n=1}^{K} k(V_m,W_n).
\end{aligned}
\label{eq:app_mmd_unbiased}
\end{equation}

Let $P_T(\cdot\mid c)$ and $P_S(\cdot\mid c)$ denote the teacher and student conditional video distributions.
We set $P=P_T$ and $Q=P_S$ for teacher--student comparisons (Table~\ref{tab:video_distribution_mmd}), and use two independent teacher sample groups with $P=Q=P_T$ for the Teacher--Teacher reference.
The latter has a theoretical $\mathrm{MMD}^2$ of zero, while finite-sample estimates can be negative.
Lower values indicate closer alignment with the teacher's conditional video distribution under the perceptual kernel.

\subsection{Evaluation Protocol}
\label{sec:app_mmd_protocol}

We evaluate all models on the same 30 conditioning histories
from 10 RoboTwin 2.0 tasks~\cite{chen2026robotwin}.
For each condition, LingBot-VA~\cite{li2026causal} generates
two independent groups of eight videos with 25 video
denoising steps and CFG scale 5.
Each student model generates eight videos with one video
denoising step and CFG scale 1.
Initial video noises are shared across student models
and independent of the teacher samples.

We compute video distances using AlexNet LPIPS over
eight predicted frames from each of three camera views.
For each condition, the bandwidth $h$ is the median
pairwise distance among the 16 teacher videos.
All teacher--student comparisons use the first teacher group
and the same bandwidth for each condition.

We average the scores within each task and then
equally across the 10 tasks.

\section{Training details}
\label{sec:supp-implementation}

\subsection{Action-aware drifting}
\label{sec:app_drifting_details}

We extract video features from the frozen teacher's 19th Transformer block at noise level $t_f=0.1$.
For two candidate videos $z,z'$, let $x=\phi^{\mv}(z)$ and $x'=\phi^{\mv}(z')$ be their extracted features.
Let $\mathcal P$ contain all unordered feature pairs from distinct candidates within the same context in the training batch, including both teacher and student candidates.
The distances used in the joint kernel are
\begin{equation}
\begin{aligned}
    D^{\mv}(x,x')
    &=
    \frac{\operatorname{RMS}(x-x')}
         {\operatorname{mean}_{(u,u')\in\mathcal P}
          \operatorname{RMS}(u-u')},\\
    D_j^{\ma}(z,z')
    &=
    \operatorname{MSE}\!\left(r_j(z),r_j(z')\right),
\end{aligned}
\end{equation}
where $r_j$ denotes the teacher's action response defined in Sec.~\ref{sec:action_aware_video_drifting}.
The visual RMS is computed over predicted-frame features, and the action MSE is computed over predicted timesteps and active action channels. We apply batch RMS normalization to the drift before forming the regression targets.

We calibrate both bandwidths before training using teacher videos, with the normalization factor in $D^{\mv}$ computed over within-context teacher pairs.
For each calibration context $c$, let $z_j^T,z_{j'}^T$ be teacher videos with features $y_j=\phi^{\mv}(z_j^T)$ and $y_{j'}=\phi^{\mv}(z_{j'}^T)$.
We set
\begin{equation}
\begin{aligned}
    h^{\mv}
    &= \operatorname{median}_{c,\,j<j'} D^{\mv}(y_j,y_{j'}),\\
    h^{\ma}
    &= \operatorname{median}_{c,\,j\ne j'} D_{j'}^{\ma}(z_j^T,z_{j'}^T).
\end{aligned}
\end{equation}
Both bandwidths remain fixed throughout training.

\subsection{Training algorithm}
\label{sec:app_algorithm}

Algorithm~\ref{alg:driftwam_training} summarizes the training loop, and Sec.~\ref{sec:app_drifting_details} specifies the normalization used in the implementation.

\begin{algorithm*}[!t]
\caption{DriftWAM Training}
\label{alg:driftwam_training}
\small
\begin{algorithmic}[1]
\Require Demonstrations $\mathcal D$, offline teacher-pair bank
$\mathcal B$, frozen teacher $\theta_T$ and video feature
extractor $\phi^{\mv}$, initial student $\theta$
\Require Student/teacher sample counts $K,M$; response noise grid
$\mathcal T_r$; CD timestep schedule
\Require Bandwidths $h^{\mv},h^{\ma}$, loss weights
$\lambda_{\mathrm{reg}},\lambda_{\mathrm{CD}},
\lambda_{\mathrm{FM}}$, EMA decay $\mu$,
point-wise pseudo-Huber distance $d$
\State $\bar\theta\gets\theta$
\For{each training iteration}
    \State $(c,z,a)\sim\mathcal D$;
    $\{(z_j^T,a_j^T)\}_{j=1}^{M}\gets\mathcal B(c)$
    \State $\epsilon_i\overset{\mathrm{i.i.d.}}{\sim}
    \mathcal N(0,I)$;
    $z_i\gets g_\theta(\epsilon_i,c)$,
    $i=1,\ldots,K$
    \State $x_i\gets\phi^{\mv}(z_i)$;
    $y_j\gets\phi^{\mv}(z_j^T)$

    \State $t\sim\operatorname{Unif}(\mathcal T_r)$;
    $\epsilon\sim\mathcal N(0,I)$;
    $a_{j,t}^T\gets(1-t)a_j^T+t\epsilon$
    \State $r_j(z')\gets
    u_{\theta_T}^{\ma}(a_{j,t}^T,t;z',c)$,
    $\quad
    z'\in\{z_i\}_{i=1}^{K}\cup\{z_j^T\},\quad j=1,\ldots,M$
    \State $k^+_{ij}\gets
    \exp\!\left[
        -\frac{\|x_i-y_j\|_2}{h^{\mv}}
        -\frac{\|r_j(z_i)-r_j(z_j^T)\|_2^2}{h^{\ma}}
    \right]$
    \State $w^+_{ij}\gets
    k^+_{ij}/\sum_{j'}k^+_{ij'}$;
    $m_i\gets\arg\max_j k^+_{ij}$
    \State $k^-_{i\ell}\gets
    \exp\!\left[
        -\frac{\|x_i-x_\ell\|_2}{h^{\mv}}
        -\frac{\|r_{m_\ell}(z_i)-r_{m_\ell}(z_\ell)\|_2^2}
        {h^{\ma}}
    \right]$,
    $\quad\ell\ne i$
    \State $w^-_{i\ell}\gets
    k^-_{i\ell}/\sum_{\ell'\ne i}k^-_{i\ell'}$
    \State $\Delta_i\gets
    \sum_j w^+_{ij}(y_j-x_i)
    -\sum_{\ell\ne i}w^-_{i\ell}(x_\ell-x_i)$
    \State $\mathcal L_{\mathrm{drift}}^{\mv}\gets
    \frac{1}{K}\sum_i
    \|x_i-\operatorname{sg}[x_i+\Delta_i]\|_2^2$
    \State $\mathcal L_{\mathrm{reg}}^{\mv}\gets
    \frac{1}{K}\sum_i
    \|r_{m_i}(z_i)-r_{m_i}(z_{m_i}^T)\|_2^2$

    \State Sample a starting timestep $t$ and obtain the target timestep $s$ from the CD schedule.
    \State $\epsilon\sim\mathcal N(0,I)$;
    $a_t\gets(1-t)a+t\epsilon$
    \State $\widetilde a_s\gets
    a_t+(s-t)u_{\theta_T}^{\ma}(a_t,t;z,c)$
    \State $\mathcal L_{\mathrm{CD}}^{\ma}\gets
    d\!\left(
        a_t-tu_\theta^{\ma}(a_t,t;z,c),\,
        \widetilde a_s
        -su_{\bar\theta}^{\ma}(\widetilde a_s,s;z,c)
    \right)$
    \State $\mathcal L_{\mathrm{FM}}^{\ma}\gets
    \|u_\theta^{\ma}(a_t,t;z,c)-(\epsilon-a)\|_2^2$

    \State $\mathcal L\gets
    \mathcal L_{\mathrm{drift}}^{\mv}
    +\lambda_{\mathrm{reg}}\mathcal L_{\mathrm{reg}}^{\mv}
    +\lambda_{\mathrm{CD}}\mathcal L_{\mathrm{CD}}^{\ma}
    +\lambda_{\mathrm{FM}}\mathcal L_{\mathrm{FM}}^{\ma}$
    \State Update $\theta$ using $\nabla_\theta\mathcal L$
    \State $\bar\theta\gets\mu\bar\theta+(1-\mu)\theta$
\EndFor
\end{algorithmic}
\end{algorithm*}

\subsection{Training configuration}
\label{sec:app_hyperparams}

On RoboTwin 2.0, we initialize the student from Flash-WAM by default and from LingBot-VA in the initialization ablation.
Both configurations use 500 optimizer steps on four NVIDIA A800 GPUs, with the hyperparameters listed in Table~\ref{tab:app_hyperparams}.
We generate the offline teacher video--action pairs with $25\mathrm{v}/50\mathrm{a}$ and video/action CFG scales of $5/1$.
The video and action noise-schedule shifts are $5$ and $1$, respectively.
We use a pseudo-Huber discrepancy for the action consistency loss.

\paragraph{Training cost.}
On RoboTwin 2.0, we measure wall-clock training time under matched settings on four NVIDIA A800 GPUs. For the same 500 optimizer steps, Flash-WAM takes about 8 hours and DriftWAM takes about 24 hours. Starting directly from LingBot-VA, however, DriftWAM surpasses the Flash-WAM baseline in average success rate under both inference budgets after just 500 steps (Table~\ref{tab:ablation_init} and Table~\ref{tab:app_ablation_init_1v2a}). This optimization run is shorter than the estimated 33 hours for Flash-WAM's 2,000-step budget~\cite{akbari2026flashwam}, extrapolated from our measured throughput. Notably, DriftWAM also incurs a one-time cost of 127 hours on the same hardware to generate the offline teacher video--action pairs, which are reused across training runs.

\begin{table}[t]
    \centering
    \small
    \caption{DriftWAM training hyperparameters on RoboTwin 2.0.}
    \label{tab:app_hyperparams}
    \begin{tabular}{@{}lc@{}}
        \toprule
        Hyperparameter & Value \\
        \midrule
        Student / teacher candidates $(K,M)$ & $(4,4)$ \\
        $\lambda_{\mathrm{reg}}$ & $0.1$ \\
        $\lambda_{\mathrm{CD}}$ & $1.0$ \\
        $\lambda_{\mathrm{FM}}$ & $0.01$ \\
        Optimizer & AdamW \\
        Learning rate & $2\times10^{-6}$ \\
        AdamW $(\beta_1,\beta_2)$ & $(0.9,0.999)$ \\
        Weight decay & $0$ \\
        Warmup steps & $100$ \\
        Global batch size & $32$ \\
        EMA decay $\mu$ & $0.995$ \\
        \bottomrule
    \end{tabular}
\end{table}

\section{Additional experiments}
\label{sec:app_more_results}

\subsection{Continued-training control under the \texorpdfstring{$1\mathrm{v}/1\mathrm{a}$}{1v/1a} setting}
\label{sec:app_continued_training}

\begin{table}[t]
  \centering
  \small
  \captionsetup{skip=8pt}
  \renewcommand{\arraystretch}{1.15}
  \caption{RoboTwin 2.0 continued-training comparison (success rates, \%) under the one-step ($1\mathrm{v}/1\mathrm{a}$) setting.
  Both Flash-WAM (+500 steps) and DriftWAM start from the official Flash-WAM checkpoint and use 500 optimizer steps.
  Each model is evaluated on 1,000 episodes (10 per task per split across 50 tasks).}
  \label{tab:continued_training}
  \setlength{\tabcolsep}{5.8pt}
  \begin{tabular*}{\columnwidth}{@{\extracolsep{\fill}}lccc}
    \toprule
    Method & Clean & Rand. & Avg. \\
    \midrule
    Flash-WAM (official) & 78.80 & 73.20 & 76.00 \\
    Flash-WAM (+500 steps) & 78.20 & 74.60 & 76.40 \\
    \textbf{DriftWAM (ours)} & \textbf{84.00} & \textbf{82.40} & \textbf{83.20} \\
    \bottomrule
  \end{tabular*}
\end{table}

On RoboTwin 2.0, our default DriftWAM student starts from the official Flash-WAM checkpoint and trains for 500 optimizer steps.
To verify that the gains come from our objective rather than extra optimization, we continue training from the same checkpoint for 500 steps with the Flash-WAM objective and evaluate all models under the one-step ($1\mathrm{v}/1\mathrm{a}$) setting on a shared 1,000-episode protocol (Table~\ref{tab:continued_training}).
Continuing Flash-WAM training yields no notable difference from the official checkpoint, suggesting that it has already largely converged under its original objective.
Under the same initialization and step budget, DriftWAM clearly outperforms the continued Flash-WAM run on both splits, confirming that its gains come from action-aware video drifting and regularization rather than additional training steps.

\subsection{Component ablation under the \texorpdfstring{$1\mathrm{v}/2\mathrm{a}$}{1v/2a} setting}
\label{sec:app_ablation_1v2a}

Table~\ref{tab:ablation_robotwin_1v2a} reports the component ablation on RoboTwin 2.0 under the $1\mathrm{v}/2\mathrm{a}$ inference budget, complementing the one-step ($1\mathrm{v}/1\mathrm{a}$) ablation in Table~\ref{tab:ablation_robotwin}.

\begin{table}[t]
  \small
  \centering
  \captionsetup{skip=8pt}
  \renewcommand{\arraystretch}{1.15}
  \caption{Ablation study of DriftWAM components on RoboTwin 2.0 (success rates, \%) under the $1\mathrm{v}/2\mathrm{a}$ setting.
  All drifting variants are initialized from Flash-WAM.
  $^*$Results reproduced by us.
  Best and second-best results are bolded and underlined, respectively.}
  \label{tab:ablation_robotwin_1v2a}
  \setlength{\tabcolsep}{5.8pt}
  \begin{tabular*}{\columnwidth}{@{\extracolsep{\fill}}lccc}
    \toprule
    Method Variant & Clean & Rand. & Avg. \\
    \midrule
    Flash-WAM$^*$ & 87.10 & 82.42 & 84.76 \\
    Vanilla Video Drifting & 87.56 & 84.12 & 85.84 \\
    Vanilla + Action-Aware Kernel & 87.46 & \underline{84.68} & 86.07 \\
    Vanilla + Action-Guided Reg. ($\mathcal{L}^{\mv}_{\mathrm{reg}}$) & \textbf{88.04} & 84.34 & \underline{86.19} \\
    \textbf{DriftWAM (Full)} & \underline{87.72} & \textbf{85.94} & \textbf{86.83} \\
    \bottomrule
  \end{tabular*}
\end{table}

Under the $1\mathrm{v}/2\mathrm{a}$ budget, vanilla video drifting improves average success over Flash-WAM. Adding either the action-aware kernel or the action-guided regularizer improves it further, and full DriftWAM achieves the highest average success rate. On the Clean split, the drifting variants perform similarly, with the variant that adds only the regularizer scoring slightly higher. One possible explanation for the narrow spread is that the second action denoising step allows the action branch to refine its predictions, making task success less sensitive to imperfections in predicted videos. On the Randomized split, adding the kernel alone gives a larger gain than adding the regularizer alone, and combining them gives the best result. The regularizer also yields a larger gain with the joint kernel than with visual-only matching, consistent with the joint kernel selecting more action-compatible pairs for regularization.

\subsection{Student initialization under the \texorpdfstring{$1\mathrm{v}/2\mathrm{a}$}{1v/2a} setting}
\label{sec:app_init_1v2a}

Table~\ref{tab:app_ablation_init_1v2a} reports the results of initializing DriftWAM directly from LingBot-VA under the $1\mathrm{v}/2\mathrm{a}$ budget.

\begin{table}[t]
  \centering
  \small
  \captionsetup{skip=8pt}
  \renewcommand{\arraystretch}{1.15}
  \caption{Ablation study on student initialization on RoboTwin 2.0 (success rates, \%) under the $1\mathrm{v}/2\mathrm{a}$ setting.
  $^*$Results reproduced by us.}
  \label{tab:app_ablation_init_1v2a}
  \setlength{\tabcolsep}{5.8pt}
  \begin{tabular*}{\columnwidth}{@{\extracolsep{\fill}}lccc}
    \toprule
    Method & Clean & Rand. & Avg. \\
    \midrule
    Flash-WAM$^*$ & 87.10 & 82.42 & 84.76 \\
    \textbf{DriftWAM (LingBot-VA init.)} & \textbf{88.06} & \textbf{83.20} & \textbf{85.63} \\
    \bottomrule
  \end{tabular*}
\end{table}

Starting from the pretrained LingBot-VA teacher, DriftWAM achieves higher success rates than Flash-WAM on both the Clean and Randomized splits.
Together with the one-step results in Table~\ref{tab:ablation_init}, this supports DriftWAM's effectiveness as a standalone distillation framework across both inference budgets.

\subsection{Inference Latency and Speedup}
\label{sec:app_latency}

\begin{table}[t]
  \small
  \centering
  \captionsetup{skip=8pt}
  \renewcommand{\arraystretch}{1.15}
  \caption{Per-chunk inference latency (ms) and speedup relative to the
  LingBot-VA teacher on a single NVIDIA GeForce RTX 4090 GPU.}
  \label{tab:app_latency}
  \setlength{\tabcolsep}{4pt}
  \begin{tabular*}{\columnwidth}{@{\extracolsep{\fill}}lccc}
    \toprule
    Inference setting & $N^{\mv}/N^{\ma}$
      & \shortstack{Mean $\pm$ Std.\\(ms)} & Speedup \\
    \midrule
    Teacher setting & $25/50$ & $4755.85 \pm 38.34$ & $1.00\times$ \\
    Distilled setting & $1/2$ & $277.25 \pm 7.32$ & $17.15\times$ \\
    Distilled setting & $1/1$ & $221.58 \pm 3.86$ & $21.46\times$ \\
    \bottomrule
  \end{tabular*}
\end{table}

\enlargethispage{2\baselineskip}
Table~\ref{tab:app_latency} reports the mean and standard deviation of the
per-chunk inference latency, together with the speedup relative
to the LingBot-VA teacher, measured on a single NVIDIA GeForce RTX 4090
GPU. Since all distilled methods share the same network architecture
and inference path, latency and speedup are reported once for each
inference setting. Here, $N^{\mv}$ and $N^{\ma}$ denote the numbers of video
and action denoising steps, respectively.

\twocolumn[{%
  \subsection{Additional Qualitative Comparisons}
  \label{sec:app_additional_qualitative}
  \begin{minipage}{\textwidth}
    \centering
    \captionsetup{type=figure}
    \includegraphics[width=\linewidth]{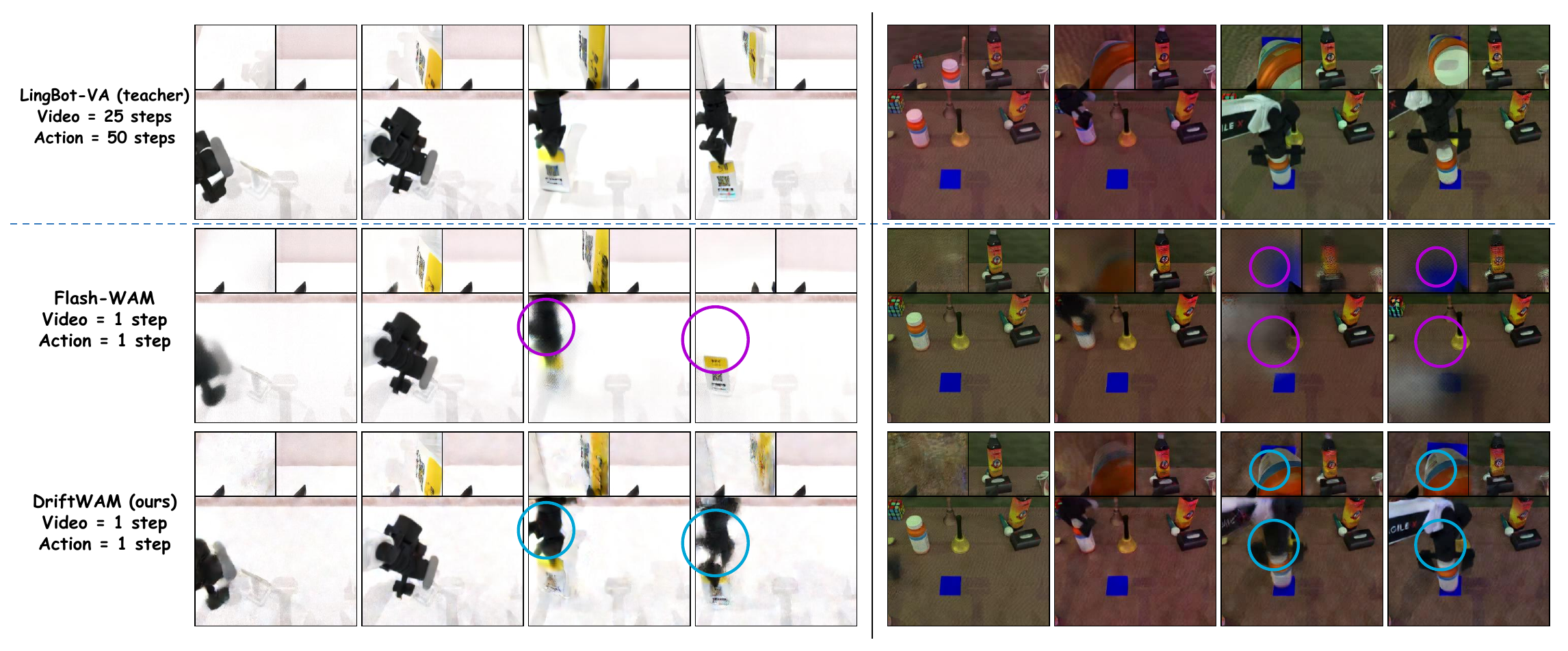}
    \caption{Additional qualitative comparisons on RoboTwin 2.0.
    Circles mark regions for comparison.}
    \label{fig:app_additional_qualitative}
  \end{minipage}\par
}]

\end{document}